\documentclass{article}
\usepackage{arxiv}
\usepackage[utf8]{inputenc}
\usepackage[english]{babel}
\usepackage{longtable}
\usepackage{placeins}
\usepackage{amsmath}
\usepackage{graphicx}
\usepackage{color}
\usepackage{geometry}
\usepackage{adjustbox}
\usepackage{mathtools}
\usepackage{longtable}
\usepackage{placeins}
\usepackage{adjustbox}
\usepackage{booktabs}
\newcommand\tab[1][1cm]{\hspace*{#1}}

\usepackage[square, numbers]{natbib}
\title{A Semi-automatic Cell Tracking Process Towards Completing the 4D Atlas of \textit{C. elegans} Development}

\author{
 Andrew Lauziere \\
  Department of Mathematics\\
  University of Maryland, College Park\\
  College Park, MD 20742 \\
  \texttt{lauziere@umd.edu} \\
   \And
 Ryan Christensen \\
  Laboratory of High Resolution Optical Imaging\\
  National Institutes of Health\\
  Bethesda, MD 20892 \\
  \texttt{ryan.christensen@nih.gov} \\
  \And
 Hari Shroff \\
  Laboratory of High Resolution Optical Imaging\\
  National Institutes of Health\\
  Bethesda, MD 20892 \\
  \texttt{hari.shroff@nih.gov}}
\date{\today}

\begin{document}

\maketitle

\begin{changemargin}{2cm}{2cm} 

    $\textbf{Abstract.}$ The nematode \textit{Caenorhabditis elegans} (\textit{C. elegans}) is used as a model organism to better understand developmental biology and neurobiology. \textit{C. elegans} features an invariant cell lineage, which has been catalogued and observed using fluorescence microscopy images. However, established methods to track cells in late-stage development fail to generalize once sporadic muscular twitching has begun. We build upon methodology which uses skin cells as fiducial markers to carry out cell tracking despite random twitching. In particular, we present a cell nucleus segmentation and tracking procedure which was integrated into a 3D rendering GUI to improve efficiency in tracking cells across late-stage development. Results on images depicting aforementioned muscle cell nuclei across three test embryos suggest the fiducial markers in conjunction with a classic tracking paradigm overcome sporadic twitching. 
    
\end{changemargin}

\section{Introduction}

\begin{figure}[!ht]
\centering
\includegraphics[width=\textwidth]{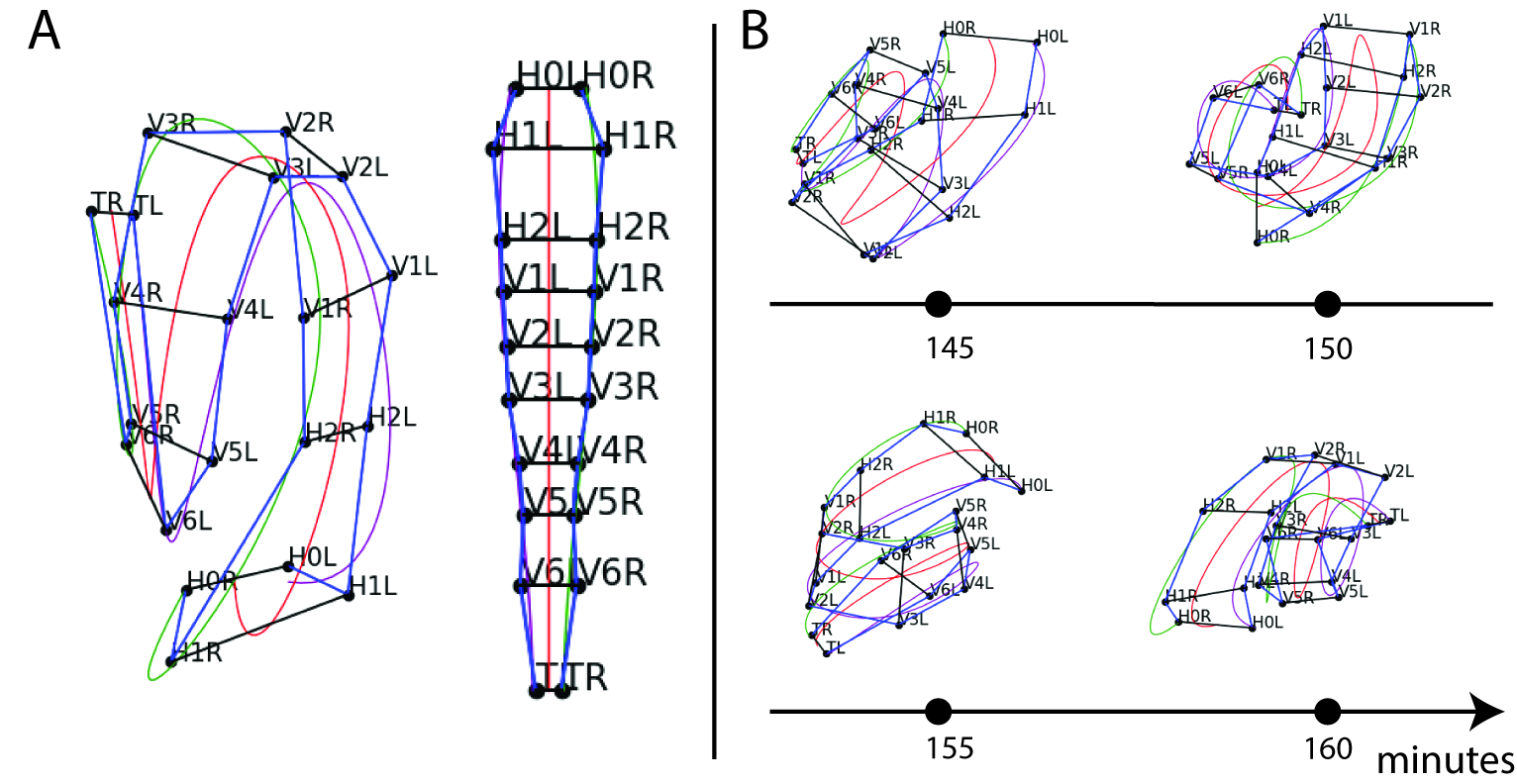}
\caption{\textbf{High spatial resolution, low temporal resolution imaging necessitates posture identification.} A: Manually identified and seam cell nuclei from an imaged \textit{C. elegans} embryo. The cells form in pairs; they are labelled posterior to anterior: \textit{T}, \textit{V6}, ..., \textit{H0}. The identification of all seam cells reveals the embryo's posture. Natural cubic splines through the left and right-side seam cells estimate the coiled body. The left image depicts identified nuclei connected to outline the embryonic worm. The fit splines are used to \textit{untwist} the worm, generating the remapped straightened points in the diagram on the right. B: Labelled nuclear coordinates from a sequence of four images. The embryo repositions in the five minute intervals between images, causing failure of traditional tracking approaches.}
\label{fig:twist_straight_3d}
\end{figure}

While the cell lineage and 4D atlas has been completed for the pre-twitch embryo, the post-twitch embryo remains a challenge \cite{santella_wormguides_2015}. The onset of muscle cells in the developing embryo causes sporadic muscular twitching. Current approaches to cell tracking in the post-twitch embryo require the identification of seam cells to approximate the coiled posture \cite{christensen_untwisting_2015}. In brief, a set of skin cells, termed \textit{seam cells} describe anatomical structure in the coiled embryo, acting as a type of ``skeleton'' outlining its body. A pair of neuroblasts appear in the final hours of development, forming an extra pair of fiducial markers. The seam cells and neuroblasts form in lateral pairs along the left and right sides of the worm, resulting in eleven pairs upon hatching \cite{sulston_embryonic_1983}. The pairs of cells are named, posterior to anterior: \textit{T}, \textit{V6}, \textit{V5}, \textit{Q} (neuroblasts), \textit{V4}, \textit{V3}, \textit{V2}, \textit{V1}, \textit{H2}, \textit{H1}, and \textit{H0}. Each pair's left and right cell is named accordingly; for example, \textit{H1L} and \textit{H1R} comprise the \textit{H1} pair.  Fig~\ref{fig:twist_straight_3d}-A depicts center points of seam cell nuclei located in an example image volume as imaged in the eggshell (left) and straightened to reveal the bilateral symmetry in seam cell locations (right). Fig~\ref{fig:twist_straight_3d}-B shows four sequential images of an embryo, five minutes between images. 

\begin{figure}[!ht]
\centering
\includegraphics[width=\textwidth]{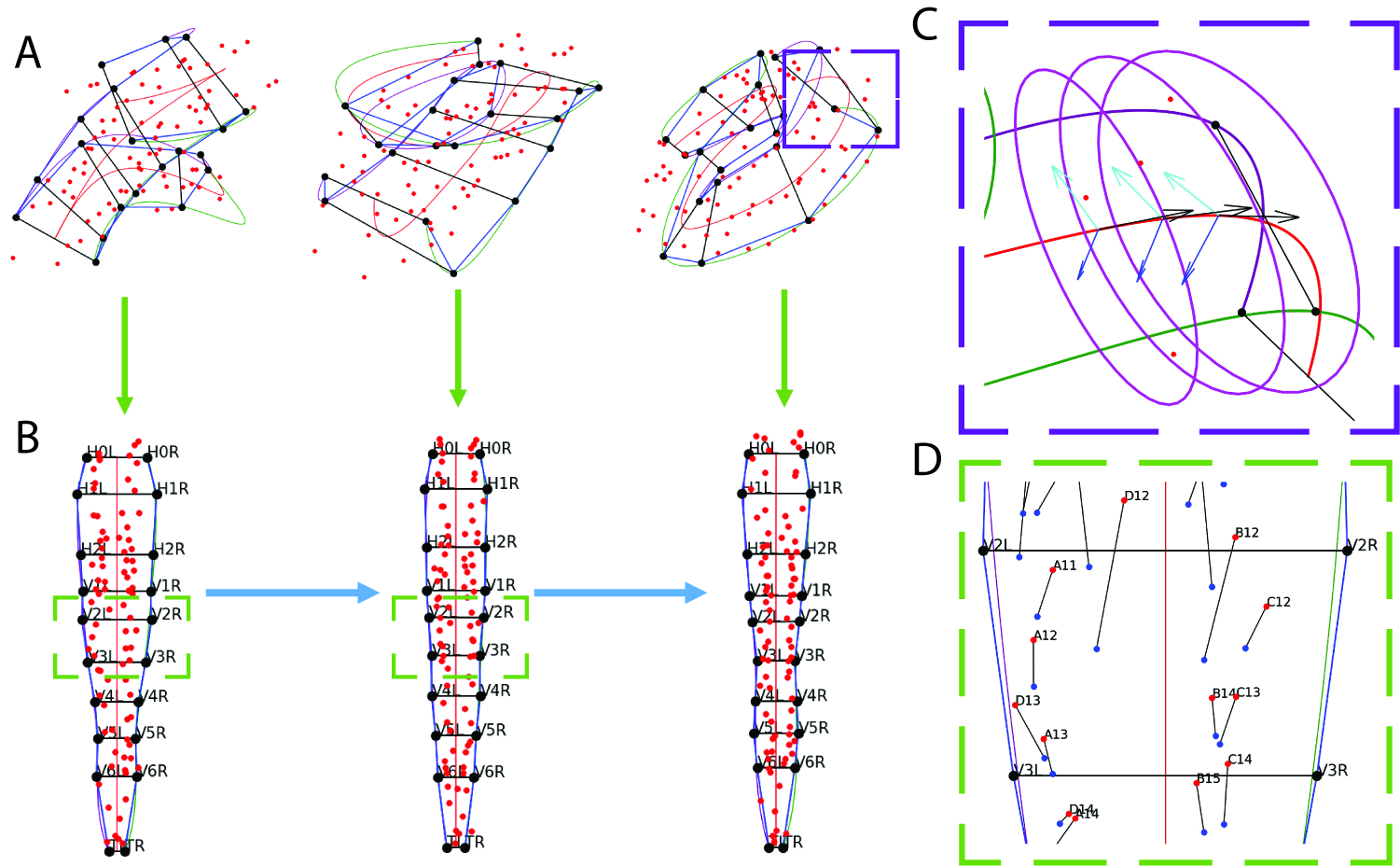}
\caption{\textbf{Posture identification allows the tracking of other cells during late-stage embrygenesis.} A: Seam cell nuclei coordinates (black) and muscle nuclei coordinates (red) in a sequence of three sequential volumetric images. The untwisting process (green arrows) uses the seam cells to remap muscle coordinates to a common frame of reference. B: The remapped muscle nuclei are tracked frame-to-frame (blue arrows). C: A higher magnification view from the right coordinate plot of A. The left, right, and midpoint splines are used to create a change of basis defined by the tangent (black), normal (blue), and binormal (cyan) vectors. Ellipses are inscribed along the tangent of the midpoint spline, approximating the skin of the coiled embryo. D: A portion of the left (red) and center (blue) remapped muscle coordinates. Black lines connect the coordinates, frame-to-frame.}
\label{fig:untwist_track}
\end{figure}

\textit{Untwisting} the worm describes the computational process of using the identified seam cell nuclei coordinates to remap coordinates of other imaged cells to a straightened coordinate system \cite{christensen_untwisting_2015}. Subsets of cells can be imaged, detected, remapped, then tracked. Fig~\ref{fig:untwist_track}-C illustrates the untwisting process on a magnified portion of the right coordinate set in Fig~\ref{fig:untwist_track}-A. The remapping uses the left, right, and midpoint splines to establish a change of basis to a straightened coordinate space defined by the tangent (black), normal (blue), and binormal (cyan) vectors. Encountered nuclei are remapped according to the position along the spline and location within the closest inscribed ellipse. Black lines connect cell nuclei in the left frame (red) to their locations in the middle frame (blue); muscle cell names arise from the four bands (A-D), anterior to posterior (Fig~\ref{fig:untwist_track}-D). 

\begin{figure}[!ht]
\centering
\includegraphics[width=\textwidth]{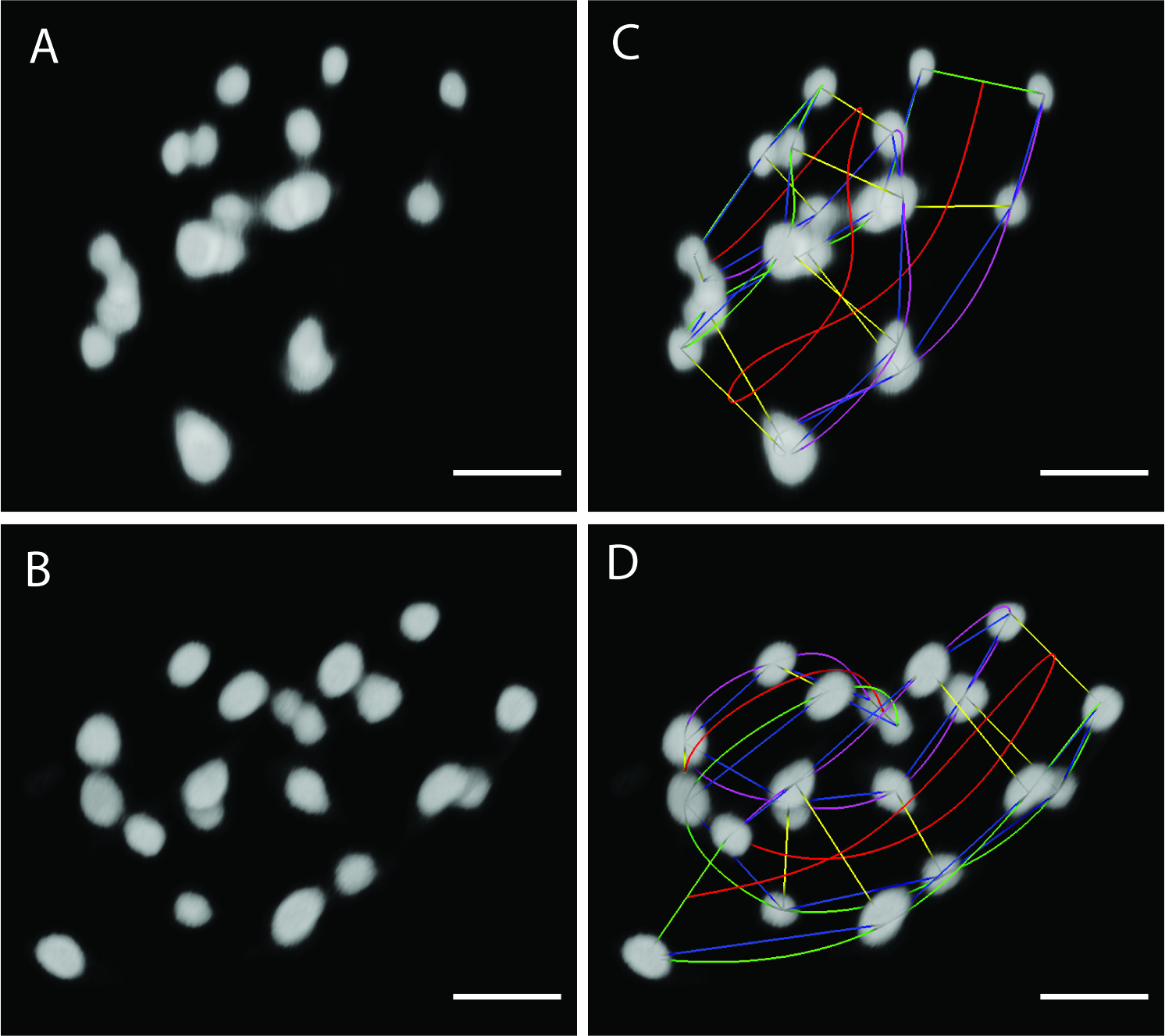}
\caption{\textbf{Manual posture identification in two successive image volumes of Fig~\ref{fig:twist_straight_3d}-B using MIPAV.} The 20 fluorescently imaged seam cell nuclei rendered in two successive image volumes. Scale bar: 10 $\mu m$. A \& B: Seam cell nuclei appearing in two successive image volumes visualized in MIPAV. The five minute interval allows the embryo to reposition between images, yielding entirely different postures. C \& D: Manual seam cell identification by trained users reveals the posture. The curved lines are cubic splines as described in Fig~\ref{fig:untwist_track}-C.}
\label{fig:model_prog-MIPAV}
\end{figure}

Current methods for seam cell identification rely on trained users to manually annotate the imaged nuclei using a 3D rendering tool \cite{mcauliffe_medical_2001}. The process takes several minutes per image volume and must be performed on approximately 100 image volumes per embryo \cite{christensen_untwisting_2015}. See \cite{lauziere_exact_2022} for a approaches to automatic seam cell identification. Fig~\ref{fig:model_prog-MIPAV} depicts manually identified seam cells in the first two successive image volumes of Fig~\ref{fig:twist_straight_3d}-B. Manual identification is performed in Medical Imaging, Processing, Analysis and Visualization (MIPAV), a 3D rendering program used for manual annotation \cite{mcauliffe_medical_2001}. 

\section{Methods}

\subsection{\textit{Untwisting} the embryo}

Consider detected cell nucleus center positions across two sequential frames. The displacement of the cell nucleus \textit{i} from frame \textit{t-1} to \textit{t} can be approximately decomposed:

\begin{equation}
    \| \mathbf{x}^{(t)}_{i} - \mathbf{x}^{(t-1)}_{i} \|_2 = \Delta D^{(t)}_{i}  \approx \Delta^I D^{(t)}_{i} + \Delta^O D^{(t)}_{i}
\end{equation}

where $\Delta^I D^{(t)}_{i}$ measures movement of nucleus \textit{i} \textit{inside} the embryo and $\Delta^O D^{(t)}_{i}$ measures nuclear movement attributed to the embryo repositioning between images. The untwisting process mitigates $\Delta^O D^{(t)}_{i}$, movement attributed to the embryo moving. The remaining term $\Delta^I D^{(t)}_{i}$ can be modeled with MOT methods. However, the untwisting process is imperfect; the seam cells represent an approximation of the coiled embryo. 

Frame-to-frame displacement of cell \textit{i} at time \textit{t}: $\Delta D^{(t)}_{i}$ is unpredictable due to the embryo repositioning between five minute images. The seam cells are used to estimate complete cubic splines through the left and right sides of the embryo \cite{christensen_untwisting_2015}. The remapped coordinates (Fig.~\ref{fig:untwist_track}) mitigate the displacement attributed to embryo movement $\Delta^O D^{(t)}_{i}$, leaving only the \textit{inter}-worm cell movement $\Delta^I D^{(t)}_{i}$. This movement can be assumed to be Brownian between frames, allowing for the application of MOT strategies to track ``straightened'' nuclei. Sample data of the 85 muscle nuclei (Fig.~\ref{fig:untwist_track}-B) are used to evaluate the efficacy of automated tracking approaches. 

\subsection{Nuclei Detection} \label{sect:DenseDetection}

Detect-and-track MOT paradigms rely to finding objects in images then associating unique objects between frames. Nuclei are detected in the sample images and compared to accompanying annotated coordinate sets. Convolutional neural networks (CNNs) achieve state of the art performance in image segmentation tasks. Two approaches are evaluated in this exploration: a 3D U-Net \cite{ronneberger_u-net_2015,cicek_3d_2016} and \textit{Stardist 3D} \cite{weigert_star-convex_2020,weigart_stardist_2021}. The 3D U-Net achieves the best results in image segmentation. Two 3D U-Net models are trained either from a random initialization or from similar data and of differing sizes to achieve voxel-level segmentations. The architecture itself remains constant, but the number of filters in each layer varies according to each model. The $L$ model has twice as many filters in each layer as $M$, which has twice as many filters as $S$. 3D U-Net models are trained to maximize the dice coefficient via the Adam optimization scheme \cite{kingma_adam_2017}. \textit{Stardist 3D} combines elements of a FCN such as the 3D U-Net but with a focus on identifying disjoint objects. Two configurations of \textit{Stardist 3D} are evaluated; the first uses a pretrained model trained on seam cell nuclei while the second is trained from a random initialization. 

\subsection{Nuclei Tracking}

The GNN filter serves as a canonical tool for MOT. The method describes a linear program (LP) in which measurements are uniquely associated to tracks. Distances between tracks and measurements are used to find a globally optimal pairing between points of the two sets. The GNN is expressed as the linear assignment problem (LAP). The assignment constraints enforce each track $i=1,2,\dots,n$ being uniquely matched to one detection $j=1,2,\dots,n$. The data association problem leverages the LAP to perform MOT. 

Define $\mathbf{Z}^{(t)} = [\mathbf{z}^{(t)}_1, \mathbf{z}^{(t)}_2, \dots, \mathbf{z}^{(t)}_n]$ as the states of each object $i=1, 2, \dots, n$ at time $t=1, 2, \dots, T$. State $\mathbf{z}^{(t)}_i$ describes the center position of object $i$ at time $t$. Similarly, define $\mathbf{O}^{(t)} = [\mathbf{o}^{(t)}_1, \mathbf{o}^{(t)}_2, \dots, \mathbf{o}^{(t)}_{m^{(t)}}]$ as the set of measurements at frame $t$, indexed $j=1, 2, \dots, m^{(t)}$, $t=1,2, \dots, T$. The detection step generates sets $\mathbf{O}^{(t)}, t = 1, 2, \dots, T$, while the association step concerns using the detections to update tracks $\mathbf{Z}^{(t)}, t=1, 2, \dots, T$. 

The GNN cost matrix $\mathbf{C} \in R^{n \times (m{(t)}+n)}$ (Eq~\ref{eqn:C}) specifies costs for associating measurements $j=1, 2, \dots, m^{(t)}$ to states $i=1, 2, \dots, n$. The matrix comprises two blocks of sizes $n \times m(t)$ and $n \times n$. The first block measures the euclidean distance between track $\mathbf{z}^{(t-1)}_i$ and detection $\mathbf{o}^{(t)}_j$ while entries in the second block consists of costs of non association known as \textit{gates}. Each gate $d^{(t)}_i$ allows for track $i$ to receive no measurements at time $t$. In the context of the GNN filter, the gate specifies a distance radius about each track $\mathbf{z}_i^{(t-1)}$ in which measurements $\mathbf{o}_j^{(t)}$ must reside in order to associate to track $i$. 

\begin{equation}
    \mathbf{C}^{(t)} = \begin{bmatrix}
\| \mathbf{z}_1^{(t-1)} - \mathbf{o}^{(t)}_1 \|_2  & \| \mathbf{z}_1^{(t-1)} - \mathbf{o}^{(t)}_2 \|_2 & ... & \| \mathbf{z}_1^{(t-1)} - \mathbf{o}^{(t)}_{m(t)} \|_2 & d_1 & \infty & \infty & ... & \infty \\
\| \mathbf{z}_2^{(t-1)} - \mathbf{o}^{(t)}_1 \|_2  & \| \mathbf{z}_2^{(t-1)} - \mathbf{o}^{(t)}_2 \|_2 & ... & \| \mathbf{z}_2^{(t-1)} - \mathbf{o}^{(t)}_{m(t)} \|_2 & \infty & d_1 & \infty & ... & \infty \\
\| \mathbf{z}_3^{(t-1)} - \mathbf{o}^{(t)}_1 \|_2  & \| \mathbf{z}_3^{(t-1)} - \mathbf{o}^{(t)}_2 \|_2 & ... & \| \mathbf{z}_3^{(t-1)} - \mathbf{o}^{(t)}_{m(t)} \|_2 & \infty & \infty & d_3 & ... & \infty \\
... & ... & ... & ... & \infty & \infty & \infty & ... & \infty \\
\| \mathbf{z}_n^{(t-1)} - \mathbf{o}^{(t)}_1 \|_2  & \| \mathbf{z}_n^{(t-1)} - \mathbf{o}^{(t)}_2 \|_2 & ... & \| \mathbf{z}_n^{(t-1)} - \mathbf{o}^{(t)}_{m(t)} \|_2 & \infty & \infty & \infty & ... & d_n \\ \end{bmatrix}  
\label{eqn:C}
\end{equation}

The cost matrix \textbf{C} defines the GNN filter objective, while the LAP constraints complete the optimization problem. The resulting linear program is then solvable in polynomial time \cite{kuhn_hungarian_1955,jonker_shortest_1987}, yielding globally optimal assignments between tracks and measurements. The columns $m(t)+1, m(t)+2, \dots, m(t)+n$ correspond to a track receiving no detection update are included in the one-to-one constraints. 

    \begin{equation}
        \begin{aligned}
                    & \text{min}
& & \sum_{i=1}^n \sum_{j=1}^{m^{(t)}} \mathbf{C}^{(t)}_{ij} x_{ij} \\
& \text{s.t.} & &  \sum_{i=1}^{n} x_{ij} = 1 \tab j = 1, \dots n+m^{(t)} \\
& & &  \sum_{j=1}^{m^{(t)}} x_{ij} = 1 \tab i = 1, 2, \dots n \\
& & &  x_{ij} \in \{0, 1\} \\
            \end{aligned}
            \label{eqn:GNN_LP}
    \end{equation} 
    
The simplest paradigm for MOT is to solve sequential frame-to-frame GNN LPs (Eq~\ref{eqn:GNN_LP}). The random motion \textit{within} the embryo is amenable to the L2 norm cost; however, the lack of trajectory to cellular motion invalidates a dynamical model such as the Kalman Filter \cite{emil_rudolph_kalman_new_1960}. However, the coordinate remapping is often imperfect and may shift nuclei from their true locations due to the spline curves not perfectly adhering to the embryo body wall. A graph can specify interactivity among adjacent nuclei, allowing for a stronger representation of nuclear movement than is possible with the GNN \cite{caetano_learning_2009,zhou_factorized_2016}. Graph matching  may provide stronger results, but the computation required due to the high number of muscle nuclei forces the use of heuristic algorithms. Kernelized Graph Matching (KerGM) \cite{zhang_kergm_2019} is the most recent development in heuristic graph matching and is applied to track straightened muscle nuclei. While GNN based solutions have been adapted for object disappearance, reappearance, merging, and splitting \cite{jaqaman_robust_2008,cox_efficient_1996}, KerGM and such graphical methods have not been adapted. Indeed, the methods can perform one-to-many or many-to-one assignment, but the methods cannot handle the common scenario in which a cell disappears in one image, and a different cell appears in the subsequent image. 

However if an LAP yields strong solutions, then evaluating multiple solutions via a more complex cost could improve the tracking quality. Murty's algorithm allows for returning the $K$ best solutions to an LAP with complexity $\mathcal{O}(Kn^3)$ \cite{murty_algorithm_1968,miller_optimizing_1997}. The $K$ hypothesized solutions to the LAP can then be evaluated on a quadratic cost. Evaluating a higher order cost is computationally inexpensive compared to searching the entire space for the assignment that minimizes the quadratic cost. If the LAP returns high quality solutions, then sampling and evaluating at the QAP cost could improve performance by further applying a more complex model to discriminate between competing hypotheses.

Both \textit{KerGM} and the evaluation of a graph cost in Murty's algorithm require a specified graph for each detection set. The choice in graph greatly influences the results. A QAP with no edge-wise connections reduces to an LAP. On the other hand, an overly dense edge set may not generalize well to unseen data. Thus several types of graphs are evaluated. The first four graphs connect nuclei in the straightened space that are withing $5 \mu m$, $7.5 \mu m$, $10 \mu m$, and $12.5 \mu m$ respectively. Delaunay Triangulation is used as another method of generating a graphical representation of the  nuclei in the straightened space at each image \cite{barber_quickhull_1996}. 

\subsection{MIPAV Interative Tracking Plugin}

Sequential detection and matching allows for accurate frame-to-frame tracking of the straightened coordinates. However, the simple matching paradigm is susceptible to large errors from false positives and false negatives. Near perfect detections are necessary for an effective frame-to-frame matching paradigm. The Untwisting plugin \cite{christensen_untwisting_2015} for MIPAV \cite{mcauliffe_medical_2001} was augmented to perform semi-automatic frame-to-frame tracking. The 3D rendering allows for interactive point identification and the import and export of such data. An initial detection set is superimposed on the volume in a 3D viewer. The user is able to perfect the detection set by removing false detections, adding missed nuclei, and splitting erroneously clumped nuclei. A first pass of the matching model links detections in the next frame to named nuclei in the prior. The user can then make corrections if necessary, and then resolve the linear program problem given user verified matched nuclei being passed as encoded constraints. This process can be repeated to completion in real time due to the polynomial complexity of solving the GNN LP \cite{jonker_shortest_1987}. 

\begin{figure}[!htb]
\centering
\includegraphics[width=\textwidth]{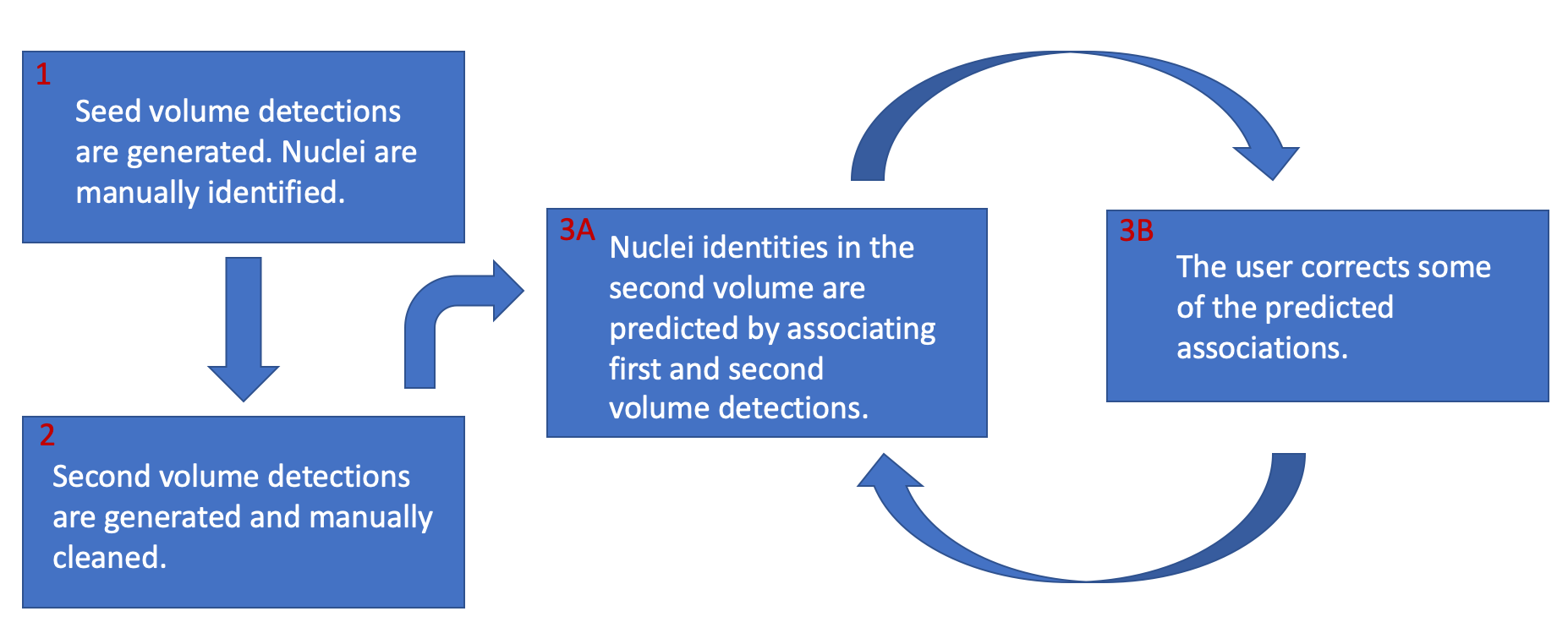}
\caption{\textbf{Iterative tuning helps overcome noisy detections.} The semi-automatic densely labelled nuclei tracking paradigm. Sets of detections in the second fluorophore are remapped via the seam cell lattice. The initial seed detections are identified (step 1). Then, the sequential frame detections are simply edited to account for all nuclei in the image (step 2). The recursive tracking procedure iteratively corrects the nuclear alignments in a manner designed to minimize the manual burden.}
\label{fig:dense_diagram}
\end{figure}

Fig~\ref{fig:dense_diagram} illustrates the semi-automated process. The seed volume detections are assessed and nuclei are identified (step 1). Then, the sequential volume detections are similarly assessed to yield an accurate detection set (step 2). Steps 3A and 3B form a recursion whereby the user generates predicted identities of the second volume nuclei (step 3A) and then edits the predicted nuclear identities. Another round of matching with added constraints will regenerate predictions (step 3B). The process is expedited with accurate detections. 

\section{Results}

A set of $n=233$ images with densely labelled cell nuclei are manually annotated to train a model to better segment smaller, closer cell nuclei. The $n$ pairs of fluorescently imaged embryos and binary annotation masks are used to fit segmentation models as described above.

\begin{figure}[!htb]
\centering
\includegraphics[width=.45\textwidth]{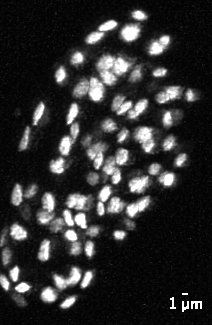}
\includegraphics[width=.45\textwidth]{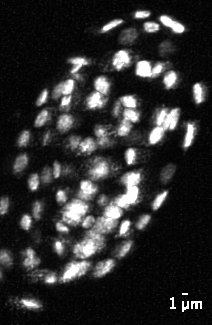}
\caption{\textbf{Densely packed muscle cell nuclei are difficult to accurately segment.} XY maximum intensity projections of two sequential image volumes depicting muscle cell nuclei. Tracking the nuclei is only feasible using seam cells as fiducial markers to remap coordinates of detected nuclei.}
\label{fig:muscle_max_proj}
\end{figure}

Images from three embryos bred to illuminate the muscle cell nuclei were annotated for evaluating the tracking methodology. The strain \textit{KP9305} targets the four bands of muscle cell nuclei within the worm embryo, each band composing approximately 21 muscle nuclei. The 85 nuclei are to the best ability of the researchers identified in each image volume throughout development for each of the three sampled worms. Nuclei centroids with associated identity are provided for all expertly detected nuclei in each image across the three worms. Figure \ref{fig:muscle_max_proj} depicts \textit{XY} maximum projections from two sequential image volumes. Cell nuclei homogeneity and density contribute to the tracking challenge.

\subsection{Detection}

Voxel-wise evaluation does not adequately measure the performance of nuclei detection models. Each model and postprocessing combination yields a set of detections to be compared to annotation volume nuclei sets. 3D U-Net models \cite{ronneberger_u-net_2015,cicek_3d_2016} are compared to \textit{Stardist 3D} \cite{weigert_star-convex_2020,weigart_stardist_2021}. Table~\ref{fig:detect_eval} highlights nuclei centroid matching results from processed image volumes in the held-out test set. \textit{Stardist 3D} is the most precise, but tends to miss more dim nuclei than the 3D U-Net models. The nuclei that can appear extremely close vary in size and intensity. Building an effective model for nuclear identification remains a barrier to automatic cell tracking.

\begin{table}[!htb]
\centering

\begin{tabular}{lrrr}
\toprule
{} &  Precision &  Recall &    F1 \\
\midrule
\textit{Stardist 3D}   &       0.81 &    0.63 &  0.68 \\
3D U-Net - S  &       0.64 &    0.67 &  0.65 \\
3D U-Net - M  &       0.69 &    0.77 &  0.72 \\
3D U-Net - L  &       0.67 &    0.75 &  0.70 \\
\bottomrule
\end{tabular}
\caption{Dense nuclei detection is more challenging than seam cell nuclei detection. The nuclei are smaller and tend to be more clustered upon each other. \textit{Stardist 3D} excels in this setting as it is the most precise model. All models struggle to find all nuclei, evidenced by the low recall measure.}
\label{fig:detect_eval}
\end{table}

\subsection{Tracking}

Manually identified nuclei positions served as test data to evaluate and compare methods. Expert annotators missed in this data, and will miss nuclei in frames due to nuclear dimming or occlusion. Tracking results arise from assuming expert level detections, i.e. that the detections are of the same quality of the annotations. Frame-to-frame accuracy is reported by the proportion of correct matches to total nuclei in the subsequent frame. Longer term frame-to-frame tracking is prone to errors, and thus tracking is currently assumed to be done in a semi-automated fashion in which a trained user edits and verifies tracks in successive frames. 

The data as presented features many frames with missing nuclei. The most challenging of scenarios occurs when two sequential frames have nuclei not present in each other. The augmented LAP is able to account for these cases by matching candidates from each set to nothing. However, the methodology is not designed for highly variant movement \textit{within} a frame \cite{jaqaman_robust_2008}. Nuclear movement introduced by untwisting error, worm elongation, and general nuclear movement are exacerbated by inaccurate detections. The GNN LP (Eq~\ref{eqn:GNN_LP}) was applied at four varying $\mu m$ gates for nuclear movement: $10 \mu m$, $15 \mu m$, $20 \mu m$, $100 \mu m$. Nuclear movement beyond each interval is a signal that a nucleus has dimmed in the subsequent frame while a newly emerging detection outside the cutoff is labelled a false positive. 

Table~\ref{fig:union_metrics} reports median matching accuracies for all gates over the test set. The gated LAP median accuracies converge to the non-gated LAP accuracy as the gate values increase. Varying the $\mu m$ cutoffs cause significant changes in the median accuracy due to highly variant movement between frames. Certain frame show low nuclear movement while others show larger elongation of the worm. The elongation \textit{stretches} coordinates along the body of the worm. Often nuclei move large distances, but they do so together. A checked detection set with an high cutoff would then allow the gated GNN LP to match nuclei despite highly variant bouts of movement.

\begin{table}[!htb]
\centering

\begin{tabular}{lrr}
\toprule
{} &  Median &    IQR \\
\midrule
\textit{GNN}                     &   0.952 &  0.120 \\
\textit{GNN} - 10 $\mu m$        &   0.615 &  0.529 \\
\textit{GNN} - 15 $\mu m$        &   0.702 &  0.564 \\
\textit{GNN} - 20 $\mu m$        &   0.708 &  0.610 \\
\textit{GNN} - 100 $\mu m$       &   0.952 &  0.145 \\
Murty - $K$ 5 - 10 $\mu m$   &   0.609 &  0.505 \\
Murty - $K$ 5 - 15 $\mu m$  &   0.690 &  0.569 \\
Murty - $K$ 5 - 20 $\mu m$   &   0.708 &  0.604 \\
Murty - $K$ 5 - 100 $\mu m$ &   0.952 &  0.134 \\
Murty - $K$ 30 - 10 $\mu m$  &   0.609 &  0.505 \\
Murty - $K$ 30 - 15 $\mu m$  &   0.701 &  0.554 \\
Murty - $K$ 30 - 20 $\mu m$  &   0.704 &  0.597 \\
Murty - $K$ 30 - 100 $\mu m$ &   0.952 &  0.135 \\
\bottomrule
\end{tabular}
\caption{The augmented LAP allows for flexibility in handling imperfect detections. However, a cardinal assumption of the method is that nuclear movement is stationary \cite{jaqaman_robust_2008}. This assumption does not hold due to the error injected from the untwisting process. A semi-automated approach in which a user corrects detections enables the \textit{GNN} model effectively match nuclei in the straightened space.}
\label{fig:union_metrics}
\end{table}

\section{Discussion}

Tracking in the remapped coordinate space using a simple \textit{GNN} tracker accurately tracks the majority of cell nuclei, but requires perfect annotations to be effective. The sources of nuclear movement are challenging to model accurately enough for reliable tracking despite the majority of frame-to-frame displacement being explained by embryo repositioning. The high variance in frame-to-frame displacement arises from error introduced by the coordinate remapping process. The body wall approximation via splines introduces systemic misplacement in some positions. The tracking approach and interface were designed to be independent of the underlying MOT approach. Future efforts to build stronger association models could be used in-place of the GNN. 

The burden thus ultimately falls on the detection process. The FCNs discussed in this work achieved results that correctly identified most of the nuclei, but are far from reliable in a fully automatic \textit{GNN} tracker. The results point to a semi-automated solution. First image volumes are processed to detect nuclei. A trained user then remove debris, splits touching nuclei, and adds annotations for missed nuclei. The \textit{GNN} tracker works in realtime, allowing for an refining approach to tracking nuclei. An initial pass will tend to correctly identify all nuclei from the prior frame. However, simply correcting any errors and rerunning while anchoring the correct predictions iteratively will produce correct frame-to-frame associations. This process applied recursively throughout an imaged worm drastically reduces both the manual effort and the time spent to track all densely labelled nuclei in a strain. 

\section{Acknowledgements}

Post-Baccalaureate research fellows Gabi Schwartz and Daniel Del Toro-Pedrosa were assisted by testing and giving feedback on the tracking interface. Alexandra Bokinksy was instrumental for the integration of the tracking backend into the MIPAV Untwisting plugin. This research is supported by the Intramural Research Program within National Institute of Biomedical Imaging and Bioengineering at the National Institutes of Health. Andrew Lauziere's contribution to this research was supported in part by NSF award DGE-1632976. This work used the computational resources of the NIH HPC Biowulf cluster. (http://hpc.nih.gov).


\end{document}